\documentclass[final,5p,times,twocolumn]{elsarticle}
\usepackage{lineno,hyperref}
\usepackage{booktabs}
\usepackage{subfigure}
\usepackage{tabularx}
\usepackage{hyperref}
\usepackage{url}
\usepackage{color, soul}
\usepackage{amsmath}
\usepackage{multirow}
\journal{Journal of \LaTeX\ Templates}

\biboptions{sort&compress}

\begin{document}

\begin{frontmatter}

\title{Multimodal Representations Learning Based on Mutual Information Maximization and Minimization and Identity Embedding for Multimodal Sentiment Analysis}

\author[mysecondaryaddress]{Jiahao Zheng}

\author[mysecondaryaddress]{Sen Zhang}

\author[mysecondaryaddress]{Xiaoping Wang\corref{mycorrespondingauthor}}
\cortext[mycorrespondingauthor]{Corresponding author} \ead{wangxiaoping@hust.edu.cn}

\author[mysecondaryaddress]{Zhigang Zeng}

\address[mysecondaryaddress]{School of Artificial Intelligence and Automation and the Key Laboratory of Image Processing and Intelligent Control of Education Ministry of China, Huazhong University of Science and Technology, Wuhan 430074, China}

\begin{abstract}
	Multimodal sentiment analysis (MSA) is a fundamental complex research problem due to the heterogeneity gap between different modalities and the ambiguity of human emotional expression. Although there have been many successful attempts to construct multimodal representations for MSA, there are still two challenges to be addressed: 1) A more robust multimodal representation needs to be constructed to bridge the heterogeneity gap and cope with the complex multimodal interactions, and 2) the contextual dynamics must be modeled effectively throughout the information flow. In this work, we propose a multimodal representation model based on Mutual information Maximization and Minimization and Identity Embedding (MMMIE). We combine mutual information maximization between modal pairs, and mutual information minimization between input data and corresponding features to mine the modal-invariant and task-related information. Furthermore, Identity Embedding is proposed to prompt the downstream network to perceive the contextual information. Experimental results on two public datasets demonstrate the effectiveness of the proposed model.
\end{abstract}


\end{frontmatter}

\section{INTRODUCTION}

\begin{figure*}[htb]
	\centering
	\includegraphics[width=160mm, height=70mm]{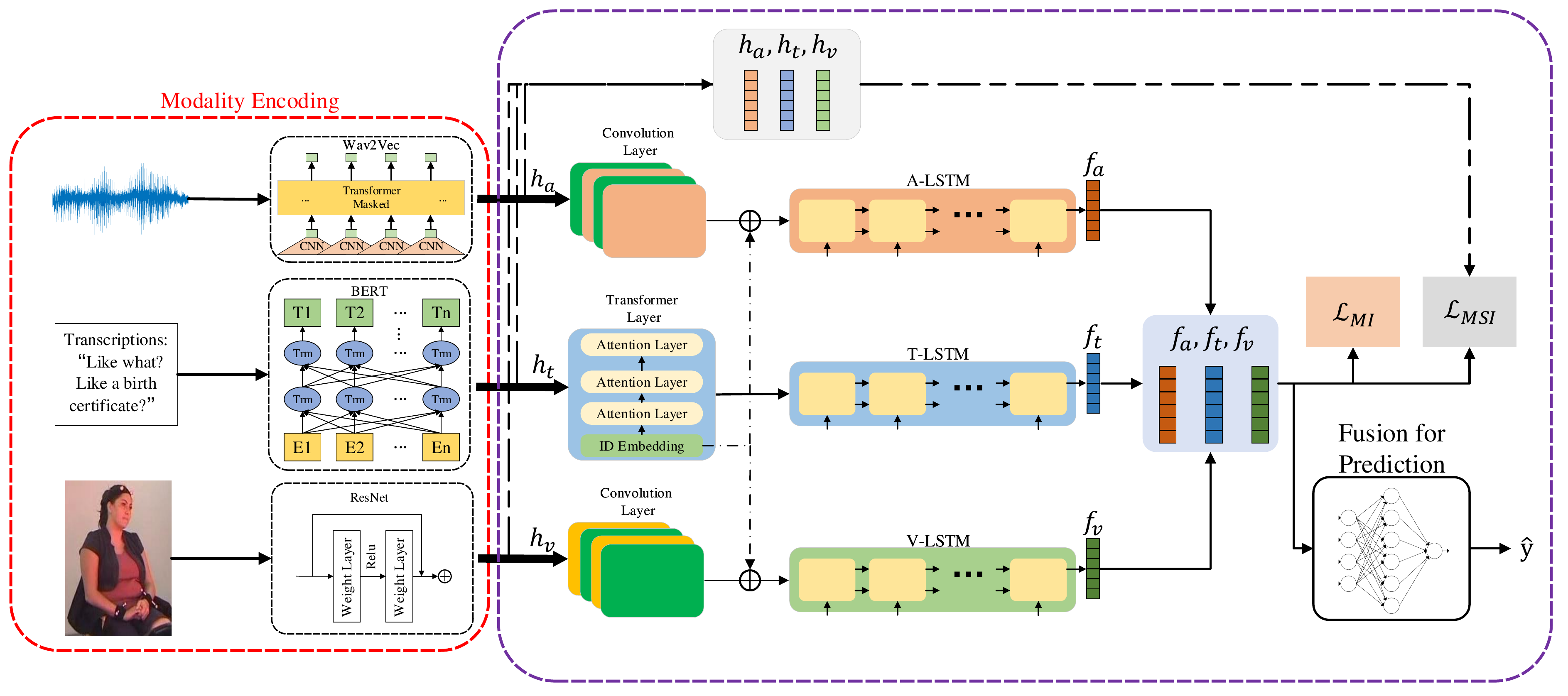}
	\caption{Illustration of the proposed MMMIE. }
	\label{fig_model}
\end{figure*}

Human perception of the world is naturally multimodal. In real life, there are three fundamental multimodal channels: text, audio and visual. The complex non-linear processing mechanism of human brain has the ability to mine modality-invariant information in these modalities, which can enhance the perception of human brain to the environments. Many literatures \cite{tsai2018learning} \cite{zadeh_multi_2018} have demonstrated that combining multimodal channels will provide additional valuable information which is beneficial to the downstream tasks in computing intelligence scenarios. However, it is difficult for the existing computer system to map the multimodal data into a unified high-dimensional space to facilitate the system decision in the multimodal scenarios. Furthermore, many of these multimodal data often contain latent emotional elements which is a reflection of a person's state of mind. Mining and recognizing these emotional elements is crucial for human-computer interaction \cite{tao2009affective}. Numerous researchers have devoted themselves to studying how to recognize the emotion category from multimodal data, and this research direction, namely multimodal sentiment analysis (MSA), has become a hot topic due to its great application potential in psychotherapy or discovering user intention.

There are two points lie at the heart of MSA: One is bridging the heterogeneity gap between multimodal information resources and constructing a common cross-modal representation space, another is modeling the contextual information in the conversation sequence. For the former one, the most common method is to utilize the strong feature extraction ability of deep neural network to map the multimodal data into a feature space and manipulate the geometric property of feature spaces. Specifically, Zadeh \textit{et al.} \cite{zadeh_tensor_2017} proposed Tensor Fusion Network (TFN) which operated a 3-fold Cartesian product between trimodal features. While TFN conducts numerous dot product operations in feature space resulting in an increase in computation. Therefore, Liu \textit{et al.} \cite{liu_efficient_2018} proposed Low-rank Multimodal Fusion (LMF) based on TFN and improved the calculation efficiency by decomposing the high-order tensors. Apart from manipulating geometric property, auxiliary loss is used to aid modal fusion. In particular, \cite{dumpala_audio_2019} proposed an autoencoder-based modal fusion method which combined reconstruction loss and canonical correlation analysis. Furthermore, Generative Adversarial Network (GAN) is also used in modal fusion. Sahu \textit{et al.} \cite{sahu_dynamic_2019} achieved cross-modal information interaction in an adversarial manner. One common characteristic of the methods mentioned above is that they construct a multimodal representation that can contain the intra-modal and cross-modal interactions. However, there are still challenges to be addressed: the multimodal representation must be robust enough to filter out the unexpected noise and the lack of overall control over information flow from input data till fusion results. The core of the latter one is utilizing the time clues in conversation sequence to improve the discrimination of the model. Early works mainly focused on how to model contextual information throughout the conversation. For instance, Hazarika \textit{et al.} \cite{hazarika2018conversational} exploited a skip attention mechanism to model the speaker's memories which appeared in the historical conversation. Recent works attempted to model contextual information from more perspectives. Specifically, Majumder \textit{et al.} \cite{majumder2019dialoguernn} modeled the emotional dynamics from current speakers, contextual content and emotion states. Ghosal \textit{et al.} \cite{ghosal2020cosmic} combined different commonsense knowledge to learn the interactions. These methods typically process features from different utterances to model the contextual information. However, this often brings a lot of extra computation and this problem becomes more serious in multimodal scenarios. Meanwhile, only focusing on the interaction between features of the deep layers will lose useful information in the shallow layers.

To solve the aforementioned intractable problems and challenges, we propose a multimodal representation model based on Mutual information Maximization and Minimization and Identity Embedding (MMMIE) to learn multimodal representations for MSA. Firstly, to achieve sufficient modal fusion throughout the information flow, Mutual Information (MI) maximization is used to mine the modal-invariant information between different modal pairs and the intractability of MI is solved by using a lower bound  estimation method based on neural network, namely MINE \cite{belghazi2018mine}. Then, to improve the robustness of the model, we introduce the concept of minimal sufficient information (MSI) which means the model should extract as little information as possible from the input data and retain the task-related information. By limiting the amount of information, the model is prompted to pay more attention to useful information and avoid the influence of noise in redundant information on performance. The minimization of MI can be solved by an upper bound estimation method, namely Contrastive Log-ratio Upper Bound (CLUB) \cite{cheng2020club}. In addition, for the contextual information, inspired by the position embedding proposed in Transformer \cite{vaswani2017attention}, we propose Identity Embedding and add it to the features of each modality, then based on attention mechanism and LSTM \cite{hochreiter1997long}, the contextual information can be modeled throughout the information flow. The effectiveness of the proposed model is demonstrated by comprehensive experiments on two large and widely used emotional datasets, i.e., the IEMOCAP \cite{busso_iemocap_2008} and the MELD \cite{poria2018meld}. Our contributions can be summarized as follows:
\begin{enumerate}[1)]
	\item We propose a fusion method that combines mutual information maximization and minimization for multimodal sentiment analysis. By maximizing the MI between modal pairs and minimizing the MI between input data and corresponding features, the effectiveness of fusion can be improved and the robustness of the multimodal representations can be ensured. 
	\item Identity embedding is proposed to model the contextual information in conversation sequence. Different from the previous work of operating on the feature space, the proposed method only needs a few trainable parameters, and benefits from the Transformer architecture, where contextual information can be prompted from shallow layers to deep layers of the network.
	\item Comprehensive experiments are designed to demonstrate the effectiveness of the proposed model. Furthermore, comparable results to the state-of-the-art models are obtained on two public datasets.
\end{enumerate}

\section{RELATED WORKS}

The proposed framework mainly focuses on a novel multimodal fusion method based on MI, and this fusion method is applied to sentiment analysis tasks. Therefore, in this section, the multimodal fusion methods are firstly overviewed. Furthermore, the related works of sentiment analysis which are the core of the proposed algorithm are reviewed in detail.

\subsection{Multimodal Fusion}
Due to the ambiguity and uncertainty of human emotions, sentiment analysis tasks based on a single modal always can not achieve satisfactory performance. Therefore, several works have employed multimodal information for sentiment analysis and demonstrated its effectiveness. They can be roughly divided into two categories, one of them is conventional fusion algorithm and the other is fusion method mainly based on some newly developed technologies. 

In the past two decades, a lot of researchers devoted themselves to studying how to bridge the heterogeneity gap among different modalities based on some conventional methods. Ngiam \textit{et al}. \cite{ngiam_multimodal_2011} proposed an autoencoder-based fusion method which mainly utilized the information interaction and unsupervised properties of the Restricted Boltzmann Machine (RBM) \cite{lee_sparse_2008}. On the basis of this approach, Wang \textit{et al}. \cite{wang_deep_2015} presented an orthogonal regularization method on the weighting matrices of the autoencoder to reduce the redundant information. The autoencoder-based fusion methods can achieve effective information interaction, but because the RBM is used as the basic module of the autoencoder, the training of these methods becomes complicated, and these methods do not have a satisfactory feature extraction capability which greatly limits their application in high-dimensional multimodal scenes. In addition to the autoencoder-based fusion method, researchers have introduced Deep Canonical Correlation Analysis (DCCA) \cite{andrew_deep_2013} into multimodal fusion tasks. Specifically, based on DCCA, literatures \cite{liu_multimodal_2019}, \cite{dumpala_audio_2019} proposed a fusion method that took the correlation between different modal features as the optimization objective and applied this method to MSA. Apart from RBM and DCCA, MI has been also utilized in modal fusion. MultiModal InfoMax (MMIM) proposed by Han \textit{et al}. \cite{han2021improving} maintained task-related information by maximizing MI in unimodal input pairs.

Besides the conventional fusion methods, some new technologies have also been introduced into the multimodal fusion tasks, such as Transformer architecture \cite{vaswani2017attention}, Generative Adversarial Networks (GAN), and contrastive learning. Tsai \textit{et al}. \cite{tsai_multimodal_2019} proposed a fusion method based on multi-head attention mechanism which was a basic module of Transformer. Through taking different modalities as Query, Key, and Value of the attention module respectively, the cross-modal information interaction was realized. Sahu \textit{et al}. \cite{sahu_dynamic_2019} proposed a GAN-based fusion method to mine the common information between different modal features by adversarial manner. Liu \textit{et al}. \cite{liu2021contrastive} proposed a method for representation learning of multimodal data using contrastive losses. Through a novel contrastive learning objective, this method could learn the complementary synergies between modalities. 

Both the conventional and the newly developed methods demonstrate that using multimodal features brings better robustness and performance than using single-modal features. This advantage is even more pronounced in sentiment analysis tasks. 

\subsection{Sentiment Analysis}
The conventional sentiment analysis tasks are always defined as identifying the emotional state of a single utterance. However, human perception of the world is based on the time domain and the expression of human emotion is temporal continuity, so only from a single utterance to recognize emotions will lose the contextual information of emotions. Therefore, the recently proposed emotion recognition works are trying to utilize contextual information of emotion to improve performance. 

In order to utilize the role of inter-speaker dependency relations, Conversational Memory Network (CMN), proposed by Hazarika \textit{et al}. \cite{hazarika2018conversational} for dyadic conversational videos, employed gated recurrent units to model the history information of each speaker into memories. Later, based on CMN, Hazarika \textit{et al}. \cite{hazarika2018icon} proposed Interactive COnversational memory Network (ICON) which modeled the inter and intra-speaker emotional influences into global memories. Based on \cite{hazarika2018conversational} \cite{hazarika2018icon}, DialogueRNN \cite{majumder2019dialoguernn} focused on using the speaker information to model emotional influence dynamically. Ghosal \textit{et al}. \cite{ghosal2019dialoguegcn} proposed DialogueGCN which introduced a graph neural network in emotion recognition to model inter and intra-dependency. Proposed by Ghosal \textit{et al}. \cite{ghosal2020cosmic}, COSMIC modeled interactions between the interlocutors within a conversation based on different elements of commonsense. Recently, Shen \textit{et al}. \cite{shen2021directed} combined the advantages of graph-based neural models and recurrence-based neural models to design a directed acyclic neural network to model the intrinsic structure of dialogue. 

The methods mentioned above demonstrate that contextual information and speaker information in dialogue are beneficial to emotion recognition. However, the performance of existing methods can be compromised when faced with problems such as lack of future information in real environments and dramatic affective variability in conversation.

\section{METHODOLOGY}

\subsection{Problem Definition}

In MSA, the input to the model is defined as a sequence of utterances $\left \{\mu_m^1, \mu_m^2, \cdots, \mu_m^T \right \}$, where $T$ is the sequence length, and $m$ represents the modality. Specifically, in this work, we have $m \in \left \{t, v, a  \right \}$, where $t, v, a$ represent textual, visual and audio modality, respectively. The core function of the designed model is to mine task-related modality-invariant information between different modalities and timing information within a sequence. Then, based on the extracted information, the model output the emotion category of each utterance in the current sequence. 

\subsection{Modality Encoding}

Firstly, the raw data needs to be encoded through the feature extraction network. Particularly, BERT \cite{devlin2018bert} is used to encode the textual modality and $h_t$ is obtained from the last hidden states of BERT. For visual modality, we use the output of fully connected layer of ResNet \cite{he_deep_2016} as visual feature and denoted as $h_v$. For audio modality, the newly developed audio recognition technology Wav2vec \cite{baevski2020wav2vec} is used. Similar to textual modality, the audio feature $h_a$ is extracted from the last hidden states of Wav2vec:
\begin{equation}
	\begin{aligned}
		h_t &= BERT(\mu_t; \theta_t) \\
		h_v &= ResNet(\mu_v; \theta_v) \\
		h_a &= Wav2vec(\mu_a; \theta_a).
	\end{aligned}
\end{equation}

\subsection{Overall Architecture}

As depicted in Fig. \ref{fig_model}, each modal raw data is firstly processed with feature extractor (firmware with no parameters to train) to get the modal features ($h_a$, $h_t$ and $h_v$). Then the features are further encoded through convolution layer or transformer layer and three different unidirectional LSTM modules are used to mine the time clues in the three encoded features. In this process, the speaker identity information is embedded into a latent space and then fed into the LSTM modules along with the three modal features. The LSTM modules output three features noted as $f_a$, $f_t$ and $f_v$. In optimization process, the model works in two collaborative parts (classification and mutual information). In the mutual information part, we exploit MINE \cite{belghazi2018mine} to maximize the MI between different modal features, and CLUB \cite{cheng2020club} is used to minimize the MI between the input features (\textit{e.g.}, $h_m, m \in \left \{t, v, a \right \}$) and output features (\textit{e.g.}, $f_m, m \in \left \{t, v, a \right \}$) within the modality. In the classification part, three modal features are fused through a multilayer perceptron. Then, the predicted categories are obtained based on the fusion results.

\subsection{Identity Embedding}

A crucial point of MSA is to model the intra- and inter-speaker dynamics. There are many trials \cite{hazarika2018icon} \cite{majumder2019dialoguernn} \cite{ghosal2019dialoguegcn} aim to propose a solid algorithm that can model these dynamics effectively. Most of them have a common characteristic, that is, they carry out geometric manipulation in the feature spaces projected by the deep neural network to model these dynamics. However, one imperfection of this manner is the inability to model information flow end to end. Identity Embedding (IE) is proposed to solve this imperfection. 

BERT which is widely used in the NLP-related tasks exploits Transformer architecture \cite{vaswani2017attention} to model the contextual information in sentences. However, due to the use of Transformer which is essentially an undirected graphical model, BERT is unable to capture the sequential information between tokens. To solve this problem, BERT exploits a position embedding which is added to the token embedding as a part of input. Then, the position information is transmitted throughout the model, from the shallow layer to the deep layer. Inspired by this, we add IE to the input. Then the dynamics in cross- and intra-speakers are modeled through the attention-based downstream network. It is represented as:
\begin{equation}
	x_i = t_i + s_i + p_i,
\end{equation}
where $x_i, i \in \left \{0, \cdots, n-1 \right \}$ is the input embedding which contains token embedding $t_i$, segment embedding $s_i$, and position embedding $p_i$. The segment embedding has a value of $0/1 $ indicates a token belongs to sentence A or sentence B. It was introduced in BERT for the next sentence prediction (NSP) task. However, recent work \cite{liu2019multi} has indicated that the NSP task is less solid. We therefore replace segment embedding with IE. Then, the $s_i$ is defined as:
\begin{equation}
	s_i = w_i \times ID_i,
\end{equation}
where the $ID_i \in \left \{ 0, 1, \cdots, T \right \}$ and $T$ represents the number of speakers in a conversation. The $w_i \in \mathbb{R}^{T \times E}$ and $E$ is the embedding dimension. It is a learnable parameter that projects the speaker ID to an embedding space. The identity information is taken as part of the input. Then, the cross- and intra-speaker dynamics can be modeled by the network based on attention mechanism in the whole information flow. The green block containing IE Embedding in Fig. \ref{fig_model} represents this module, and the dash lines represent the IE is added to other two modalities. It should be noted that the IE module needs that the identity information of the speakers in the conversation is known. In the application scenarios of emotion robots, there always exists two participants in the conversation, one is the agent and the other is the user. Therefore, we can use ``0'' to represent the agent and ``1'' to represent the user. For other application scenarios with unknown identity information, this module can be used in conjunction with the voiceprint recognition algorithm \cite{doddington1985speaker} or can be removed flexibly. Furthermore, the effectiveness of the IE module will be analyzed in the experimental section.

\subsection{Multimodal Fusion Based on MI}
Apart from modeling the dynamics between speakers, it is also necessary to model the dynamics between multi-modalities. In probability theory and information theory, the MI of two random variables is a measure of the mutual dependence between the two variables. In this work, a MI-based multimodal fusion method is proposed. Mathematically, for a pair of continuous random variables $X$ and $Y$, the MI can be defined as:

\begin{equation}
	I(X;Y) = \iint_{}^{}p(x, y) \log \frac{p(x, y)}{p(x)p(y)}dxdy,
	\label{MI_distribution}
\end{equation}
where $p(x, y)$ is the joint distribution and the marginal distributions are $p(x)$ and $p(y)$. It is equivalent to the Kullback-Leibler (KL-) divergence:
\begin{equation}
	I(X;Y) = D_{KL}(\mathbb{P}_{XY} \parallel  \mathbb{P}_X \otimes  \mathbb{P}_Y),
	\label{MI_KL}
\end{equation}
where $\mathbb{P_{XY}}$ is the joint distribution and $\mathbb{P}_X \otimes \mathbb{P}_Y$ is the product of the marginals, and $D_{KL}$ is defined as:
\begin{equation}
	D_{KL}(\mathbb{P} \otimes \mathbb{Q}) := \mathbb{E}_{\mathbb{P}} \left [ \log {\frac{d \mathbb{P}}{d \mathbb{Q}}} \right ].
	\label{KL_def}
\end{equation}

The intuitive interpretation of KL divergence is the gap between two probability distributions. Naturally, the meaning of MI defined in Eq. (\ref{MI_KL}) can be understood as: the smaller the distance between the joint distribution and the marginal distributions, the stronger the dependency between $X$ and $Y$. Due to this property of measuring dependency between random variables, MI has been widely utilized as an optimization objective in multimodal fusion or cross-modal retrieval in recent works \cite{han2021improving} \cite{zhuang2019investigation} \cite{datta2017multimodal}. 

According to Eq. (\ref{MI_distribution}), it can be seen intuitively that calculating MI requires the joint distribution and the marginal distribution are known. However, if MI is exploited for modal fusion, the input data tends to be highly dimensional and continuous, and the available samples (the datasets) are too sparse to accurately estimate the distributions in the MSA tasks. Videlicet, we can only get the posterior distribution $\mathbb{P}_{(Y|X)}$ through an encoder model based on neural network. Whereas $\mathbb{P}_{XY}$, $\mathbb{P}_X$ and $\mathbb{P}_Y$ are all intractable. Therefore, there is a gap between the empirical distributions estimated by samples and the actual distributions. 

From the above analysis, it is almost impossible to directly estimate MI in high-dimensional spaces. Therefore, we exploit MINE \cite{belghazi2018mine} which is based on neural network to obtain an accurate and tight lower bound of MI. Through MINE to maximize MI between different modal features, the cross-modality dynamics which is equivalent to modality-invariant information can be kept and the modality-specific information which is considered to be task-independent noise can be filtered out. Specifically, the lower bound of MI can be obtained through the dual representation of KL divergence:
\begin{equation}
	D_{K L}(\mathbb{P} \parallel \mathbb{Q})=\sup _{T: \Omega \rightarrow \mathbb{R}} \mathbb{E}_{\mathbb{P}}[T]-\log \left(\mathbb{E}_{\mathbb{Q}}\left[e^{T}\right]\right),
\end{equation}
where $\mathbb{P}$ and $\mathbb{Q}$ are two arbitrary distributions and $T$ is an arbitrary function mapping from the sample space $\Omega$ to the real number $\mathbb{R}$. Let $\mathcal{F}$ be any class of functions $T: \Omega \rightarrow \mathbb{R}$, then, the lower bound can be represented as:
\begin{equation}
	D_{K L}(\mathbb{P} \parallel \mathbb{Q}) \geqslant \sup _{T \in \mathcal{F}} \mathbb{E}_{\mathbb{P}}[T]-\log \left(\mathbb{E}_{\mathbb{Q}}\left[e^{T}\right]\right).
	\label{KL_dual}
\end{equation}

Combining Eq. (\ref{MI_KL}) and Eq. (\ref{KL_dual}), the lower bound of MI can be represented as:
\begin{equation}
	\begin{split}
		I(X;Y) &= D_{KL}(\mathbb{P_{XY}} \parallel \mathbb{P_X} \otimes \mathbb{P}_Y) \\
		&\geqslant \sup _{T \in \mathcal{F}} \mathbb{E}_{\mathbb{P}_{XY}}[T]-\log \left(\mathbb{E}_{\mathbb{P}_X \otimes \mathbb{P}_Y}\left[e^{T}\right]\right),
	\end{split}
\end{equation}
where $\mathcal{F}$ is a family of functions $T: \mathcal{X} \times \mathcal{Y} \rightarrow \mathbb{R}$.
Due to the neural networks can be viewed as a complex nonlinear function, we can utilize this property to replace function $T$ with a deep neural network with parameters $\theta \in \Theta$ and mark it as $T_{\theta}$:
\begin{equation}
	I_{\Theta}(X, Y) = \sup _{\theta \in \Theta} \mathbb{E}_{\mathbb{P}_{XY}}[T_{\theta}]-\log \left(\mathbb{E}_{\mathbb{P}_X \otimes \mathbb{P}_Y}\left[e^{T_{\theta}}\right]\right).
	\label{MI_theta}
\end{equation}

Furthermore, through sampling from the dataset, $\mathbb{P}_{XY}$, $\mathbb{P}_X$ and $\mathbb{P}_Y$ in Eq. (\ref{MI_theta}) can be represented by empirical distribution. We mark them as $\hat{\mathbb{P}}_{XY}^{(n)}$, $\hat{\mathbb{P}}_{X}^{(n)}$ and $\hat{\mathbb{P}}_{Y}^{(n)}$. Then, when batch size is $n$, the ultimate objective of maximizing MI can be expressed as:
\begin{equation}
	\widehat{I(X ; Y)}_{n}=\sup _{\theta \in \Theta} \mathbb{E}_{\hat{\mathbb{P}}_{X Y}^{(n)}}\left[T_{\theta}\right]-\log \left(\mathbb{E}_{\hat{\mathbb{P}}_{X}^{(n)} \otimes \hat{\mathbb{P}}_{Y}^{(n)}}\left[e^{T_{\theta}}\right]\right).
	\label{MI_objective}
\end{equation}

A more intuitive understanding of Eq. (\ref{MI_objective}) is that the lower bound of MI can be continuously improved through adjusting the nonlinear function $T_{\theta}$ (the deep neural network) based on the back propagation algorithm in the training process, and finally the MI maximization can be realized. This process is equivalent to variational approximation. Whether MI is calculated based on Eq. (\ref{MI_distribution}) or Eq. (\ref{MI_objective}), they both need to estimate the overall mathematical characteristics based on samples. The key difference, however, is that Eq. (\ref{MI_distribution}) needs to replace overall distribution with empirical distribution and Eq. (\ref{MI_objective}) needs to compute sample expectation instead of overall expectation. Calculating empirical distribution is more intractable than calculating sample expectation.

Specifically, in the multimodal fusion task, the loss function for MI maximization is:
\begin{equation}
	\mathcal{L}_{MI} = -\widehat{I(f_t ; f_v)}_{n} - \widehat{I(f_t ; f_a)}_{n} - \widehat{I(f_v ; f_a)}_{n}.
	\label{equa_mi}
\end{equation}

In the sentiment analysis tasks, the performance of data-driven model is easily affected by the data noise. Aforementioned MINE-based modal fusion method can filter the data noise by extracting modality-invariant contents. In order to further improve the robustness of the proposed model, we propose another optimization objective based on minimal sufficient information (MSI) theory. Based on MSI, the reason why the model is sensitive to data disturbance is that it lacks the ability to filter some redundant information that is not directly related to task. Referring to the work of Wu \textit{et al}. \cite{wu2020graph}, this problem can be solved by reducing the MI between the input data and the features of the model output. 

We exploit a novel MI upper bound estimator named CLUB \cite{cheng2020club} to reduce the MI. For samples $\left \{x_i, y_i \right \}$ drawn from an intractable distribution $p(x, y)$, the unbiased estimator CLUB is defined as:
\begin{equation}
	\hat{I}_{CLUB}=\frac{1}{N^{2}} \sum_{i=1}^{N} \sum_{j=1}^{N}\left[\log p\left(y_{i} \mid x_{i}\right)-\log p\left(y_{j} \mid x_{i}\right)\right],
\end{equation}
where $N$ is the batch size. In our task and the vast majority of machine learning tasks, the conditional distribution $p(y_i \mid x_i)$ is always unknown. Therefore, a variational distribution $q_{\theta}(y_i \mid x_i)$ which is realized by a deep neural network with parameters $\theta$ is utilized to approximate the real conditional distribution. Then, the variational CLUB $\hat{I}_{vCLUB}$ is defined by:
\begin{equation}
	\hat{I}_{vCLUB}=\frac{1}{N^{2}} \sum_{i=1}^{N} \sum_{j=1}^{N}\left[\log q_{\theta}\left(y_{i} \mid x_{i}\right)-\log q_{\theta}\left(y_{j} \mid x_{i}\right)\right].
\end{equation}

Having defined the MI upper bound estimator $\hat{I}_{vCLUB}$, we can get the MSI optimization objective:
\begin{equation}
	\hat{I}_{MSI}=\frac{1}{N^{2}} \sum_{i=1}^{N} \sum_{j=1}^{N}\left[\log q_{\theta}\left(f_{i} \mid h_{i}\right)-\log q_{\theta}\left(f_{j} \mid h_{i}\right)\right],
\end{equation} 
where $h_i$ is the features extracted from the pretrained network (\textit{e.g.}, $h_t$, $h_v$ and $h_a$ mentioned above) and $f_i$ is the features output by the LSTM modules (\textit{e.g.}, $f_t$, $f_v$ and $f_a$). The multimodal MSI optimization objective can be represented as:
\begin{equation}
	\mathcal{L}_{MSI} = \hat{I}_{tMSI} + \hat{I}_{vMSI}  + \hat{I}_{aMSI} ,
	\label{equa_club}
\end{equation}
where $\hat{I}_{tMSI}$, $\hat{I}_{vMSI}$ and $\hat{I}_{aMSI}$ represent the MI upper bound approximation calculated from the features of text, visual and audio modalities, respectively.

Besides the modal fusion optimization objective, we have the prediction optimization objective based on the final prediction $\hat{y}$ and the truth value $y$:
\begin{equation}
  \mathcal{L}_{task} = \sum_{i}^{N} (\hat{y}_i) \log{y_i}.
\end{equation}

Finally, the main loss can be calculated by summing up the above-mentioned losses:
\begin{equation}
	\mathcal{L}_{main} = \mathcal{L}_{task} + \alpha \mathcal{L}_{MI} + \beta \mathcal{L}_{MSI}, 
\end{equation}
where $\alpha$ and $\beta$ are both hyperparameters.



\section{EXPERIMENTS}
In this section, we introduce some experiments details, including datasets, baselines, model configuration, evaluation measures, and results.  

\subsection{Datasets and Metrics}
This work evaluates the performance of the proposed algorithm on two different datasets which are widely used in MSA research: IEMOCAP \cite{busso_iemocap_2008} and MELD \cite{poria2018meld}. 

\emph{IEMOCAP:} This dataset contains 10039 utterances in which ten professional actors (five males and five females) are expressing their emotion based on scripts or improvisation. 
The emotions (i.e., angry, happy, sad, neutral, frustrated, excited, fearful, surprised, disgusted, and others) of each participant in the utterance are manually annotated by three different annotators and the participant. Following the prior study \cite{ghosal2019dialoguegcn}, six emotions are used in this work: happy, sad, neutral, angry, excited, and frustrated.


\emph{MELD:} This dataset has more than 1400 dialogues and 13000 utterances from Friends TV series. The dataset is split to train set, validation set and test set which contains 10643, 2384 and 4361 utterances, respectively. It contains multi-party conversations that is different from the dyadic conversation of IEMOCAP. Each utterance is labeled by any of these seven emotions: anger, disgust, sadness, joy, neutral, surprise and fear. 

\emph{Metrics:} Due to IEMOCAP does not provide a standard test set, following the previous work \cite{ghosal2019dialoguegcn}, we split this dataset by session and perform leave-one-speaker-out experiments. To accord with the previous work, six-class classification accuracy is adopted as metrics set. Since the datasets used in this work is class imbalance, the weighted F1-score is also adopted.

\subsection{Baselines}
For a comprehensive evaluation of the proposed model, we compare it with many baselines. We consider some conventional works: TFN \cite{zadeh_tensor_2017}, KET \cite{zhong2019knowledge}, DialogueRNN \cite{majumder2019dialoguernn}, DialogueGCN \cite{ghosal2019dialoguegcn}, some newly proposed methods are also considered: COSMIC \cite{ghosal2020cosmic}, BiERU \cite{li2022bieru}, HiTrans \cite{li2020hitrans}, RGAT \cite{Ishiwatari2020RGAT}. The details of these baselines are listed as follows:

\textbf{TFN}: Tensor Fusion Network achieves modal fusion in feature space by using a 3-fold Cartesian product to model the unimodal, bimodal and trimodal interactions.

\textbf{KET}: Knowledge Enriched Transformer models the context by a hierarchical self-attention module and the external commonsense knowledge is dynamically introduced in the information flow through a context-aware affective graph attention mechanism.

\textbf{DialogueRNN}: This method utilizes a GRU to track the participant emotional states throughout the conversation and uses another GRU to track the global context. In general, it focuses on the inter-speaker dynamics.

\textbf{DialogueGCN}: Dialogue Graph Convolutional Network is a graph-based model where the nodes represent individual utterances and the edges represent the dependency between the speakers. Then the contextual information can be propagated to distant utterances.




\textbf{COSMIC}: This framework is commonsense-guided and captures the emotional dynamics in conversation based on a large commonsense knowledge base. 

\textbf{BiERU}: Bidirectional Emotional Recurrent Unit is a parameter-efficient party-ignorant framework. It exploits a generalized neural tensor block and a two-channel feature extractor to capture the contextual information.

\textbf{HiTrans}: This framework consists of two hierarchical transformers. One is used as a low-level feature extractor to capture the local associations and the low-level feature is fed into the another transformer to model contextual information.

\textbf{RGAT}: This framework utilizes relational graph attention networks to model self- and inter-speaker dependencies. In addition, relational position encoding is proposed to provide sequential information.


\subsection{Basic Settings and Results}

\begin{table}[htbp]
	\centering
	\caption{The hyperparameters used in the experiment. In this table, the ``lr'' and ``rs'' represent learning rate and random seed, respectively.}
	\resizebox{.5\textwidth}{!}{ 
	  \begin{tabular}{cccccccc}
			&       &       &       &       &       &       &  \\
  \cmidrule{2-7}          & \multicolumn{1}{c|}{\multirow{2}[4]{*}{Dataset}} & \multicolumn{5}{c}{Hyperparameters}   &  \\
  \cmidrule{3-7}          & \multicolumn{1}{c|}{} & batch size & $\alpha$     & $\beta$     & lr & rs &  \\
  \cmidrule{2-7}          & IEMOCAP & 2     & 0.3   & 0.0002 & 2e-5  & 100   &  \\
			& MELD  & 4     & 0.2   & 0.0006 & 4e-5     & 100   &  \\
  \cmidrule{2-7}          &       &       &       &       &       &       &  \\
	  \end{tabular}}
	\label{tab_hyperparameters}
\end{table}

\begin{table}[htbp]
	\centering
	\caption{Comparison of the proposed model with the eight baselines methods on IEMOCAP and MELD datasets. The best results are marked in bold.}
	  \begin{tabular}{ccccccc}
			&       &       &       &       &       &  \\
  \cmidrule{2-6}          & \multicolumn{1}{c|}{\multirow{3}[4]{*}{Models}} & \multicolumn{2}{c|}{MELD} & \multicolumn{2}{c}{IEMOCAP} &  \\
  \cmidrule{3-6}          & \multicolumn{1}{c|}{} & \multirow{2}[2]{*}{ACC-7} & \multicolumn{1}{p{4.19em}|}{weighted} & \multirow{2}[2]{*}{ACC-6} & \multicolumn{1}{l}{weighted} &  \\
			& \multicolumn{1}{c|}{} &       & \multicolumn{1}{c|}{F1} &       & \multicolumn{1}{c}{F1}    &  \\
  \cmidrule{2-6}          & \multicolumn{1}{c|}{TFN} & -     & \multicolumn{1}{c|}{-} & 58.80  & 58.50  &  \\
			& \multicolumn{1}{c|}{KET} & -     & \multicolumn{1}{c|}{58.18} & -     & 59.56 &  \\
			& \multicolumn{1}{c|}{DialogueRNN} & 59.54 & \multicolumn{1}{c|}{57.03} & 63.40  & 62.75 &  \\
			& \multicolumn{1}{c|}{DialogueGCN} & 59.46 & \multicolumn{1}{c|}{58.10} & 65.25 & 64.18 &  \\
			& \multicolumn{1}{c|}{COSMIC} & -     & \multicolumn{1}{c|}{\textbf{65.21}} & -     & 65.28 &  \\
			& \multicolumn{1}{c|}{BiERU} & 60.90  & \multicolumn{1}{c|}{-} & 66.09 & 64.59 &  \\
			& \multicolumn{1}{c|}{HiTrans} & -     & \multicolumn{1}{c|}{61.94} & -     & 64.50  &  \\
			& \multicolumn{1}{c|}{RGAT} & -     & \multicolumn{1}{c|}{60.91} & -     & 65.22 &  \\
  \cmidrule{2-6}          & \multicolumn{1}{c|}{MMMIE} & \textbf{65.06} & \multicolumn{1}{c|}{64.12} & \textbf{67.78} & \textbf{67.53} &  \\
  \cmidrule{2-6}          &       &       &       &       &       &  \\
	  \end{tabular}%
	\label{compare_baseline}%
\end{table}%

\begin{table*}[htbp]
	\centering
	\caption{Performance comparisons for each category on IEMOCAP. The best results are marked in bold. $\dag$ represents the results from \cite{hazarika2018icon}.}
	  \begin{tabular}{ccccccccccccccccc}
			&       &       &       &       &       &       &       &       &       &       &       &       &       &       &       &  \\
  \cmidrule{2-16}          & \multicolumn{1}{c|}{\multirow{3}[4]{*}{Models}} & \multicolumn{14}{c}{IEMOCAP: Emotion Categories}                                                              &  \\
			& \multicolumn{1}{c|}{} & \multicolumn{2}{c}{Happy} & \multicolumn{2}{c}{Sad} & \multicolumn{2}{c}{Neutral} & \multicolumn{2}{c}{Angry} & \multicolumn{2}{c}{Excited} & \multicolumn{2}{c}{Frustrated} & \multicolumn{2}{c}{\textit{\textbf{Avg.}}} &  \\
  \cmidrule{3-16}          & \multicolumn{1}{c|}{} & acc.  & \multicolumn{1}{c|}{F1} & acc.  & \multicolumn{1}{c|}{F1} & acc.  & \multicolumn{1}{c|}{F1} & acc.  & \multicolumn{1}{c|}{F1} & acc.  & \multicolumn{1}{c|}{F1} & acc.  & \multicolumn{1}{c|}{F1} & acc.  & F1    &  \\
  \cmidrule{2-16}          & \multicolumn{1}{c|}{$\text{cLSTM}^{\dag}$ \cite{poria2017context}} & 25.5  & \multicolumn{1}{c|}{35.6} & 58.6  & \multicolumn{1}{c|}{69.2} & 56.5  & \multicolumn{1}{c|}{53.5} & 70.0 & \multicolumn{1}{c|}{66.3} & 58.8  & \multicolumn{1}{c|}{61.1} & 67.4  & \multicolumn{1}{c|}{62.4} & 59.8  & 59.0  &  \\
			& \multicolumn{1}{c|}{$\text{TFN}^{\dag}$ \cite{zadeh_tensor_2017}} & 23.2  & \multicolumn{1}{c|}{33.7} & 58.0  & \multicolumn{1}{c|}{68.6} & 56.6  & \multicolumn{1}{c|}{55.1} & 69.1  & \multicolumn{1}{c|}{64.2} & 63.1  & \multicolumn{1}{c|}{62.4} & 65.5  & \multicolumn{1}{c|}{61.2} & 58.8  & 58.5  &  \\
			& \multicolumn{1}{c|}{$\text{MFN}^{\dag}$ \cite{zadeh_memory_2018}} & 24.0  & \multicolumn{1}{c|}{34.1} & 65.6  & \multicolumn{1}{c|}{70.5} & 55.5  & \multicolumn{1}{c|}{52.1} & 72.3  & \multicolumn{1}{c|}{66.8} & 64.3  & \multicolumn{1}{c|}{62.1} & 67.9  & \multicolumn{1}{c|}{62.5} & 60.1  & 59.9  &  \\
			& \multicolumn{1}{c|}{$\text{CMN}^{\dag}$ \cite{hazarika2018conversational}} & 25.7  & \multicolumn{1}{c|}{32.6} & 66.5  & \multicolumn{1}{c|}{72.9} & 53.9  & \multicolumn{1}{c|}{56.2} & 67.6  & \multicolumn{1}{c|}{64.6} & 69.9  & \multicolumn{1}{c|}{67.9} & 71.7  & \multicolumn{1}{c|}{63.1} & 61.9  & 61.4  &  \\
			& \multicolumn{1}{c|}{$\text{ICON}^{\dag}$ \cite{hazarika2018icon}} & 23.6  & \multicolumn{1}{c|}{32.8} & 70.6  & \multicolumn{1}{c|}{74.4} & 59.9  & \multicolumn{1}{c|}{60.6} & 68.2  & \multicolumn{1}{c|}{\textbf{68.2}} & 72.2  & \multicolumn{1}{c|}{68.4} & \textbf{71.9} & \multicolumn{1}{c|}{\textbf{66.2}} & 64.0  & 63.5  &  \\
  \cmidrule{2-16}          & \multicolumn{1}{c|}{MMMIE} & \textbf{59.0} & \multicolumn{1}{c|}{\textbf{56.5}} & \textbf{88.2} & \multicolumn{1}{c|}{\textbf{77.7}} & \textbf{59.9} & \multicolumn{1}{c|}{\textbf{63.2}} & \textbf{72.4} & \multicolumn{1}{c|}{66.0} & \textbf{73.9} & \multicolumn{1}{c|}{\textbf{74.9}} & 59.1  & \multicolumn{1}{c|}{64.5} & \textbf{67.8} & \textbf{67.5} &  \\
  \cmidrule{2-16}          &       &       &       &       &       &       &       &       &       &       &       &       &       &       &       &  \\
	  \end{tabular}%
	\label{category_performance}%
\end{table*}%

\textbf{Model Configuration}: We train and evaluate the proposed model on the Pytorch 1.7.0 framework with a single NVIDIA TITAN Xp GPU. In the training process, the maximum number of epoch is set as 50 and the evaluation is performed after each training epoch. If the current evaluation accuracy is better than the saved-best one, the current model parameters will be saved and the accuracy will also be updated. Limited by the computing resources, APEX-based half precision is used to reduce GPU memory usage. In the process of feature extraction, BERT with a hidden size of 768 is used to extract context information of unimodal textual modality, Wav2vec model with the same hidden size as BERT is used to extract audio features, and a 2048-dimensional visual feature is extracted by ResNet. The best set of hyperparameters is determined by running grid search. Finally, the hyperparameters used in the experiment are shown in Table \ref{tab_hyperparameters}. We fix the random seed as 100 to ensure reproducibility.

\textbf{Results.} The proposed model is compared with the baselines to demonstrate the effectiveness of the model. The experimental results are shown in Table \ref{compare_baseline}. It is easy to see the proposed model achieve better or comparable performance to many baselines. To elaborate, on IEMOCAP, the proposed model significantly outperforms the mentioned eight baselines in both accuracy and F1 metrics. On MELD, the performance of the proposed model is also second only to COSMIC. Besides the overall comparison, the performance in each category is presented in Table \ref{category_performance}.  As can be seen, MMMIE outperforms all the compared models except for anger and frustration emotion. All the methods presented in Table \ref{category_performance} achieve poor performance on happiness emotion, which is due to the small number of data samples in this category. This reflects the dependency of the model on the amount of data, which is what we need to solve in the future. It should be noted that the results of the proposed model in Table \ref{category_performance} is calculated by sklearn\footnote{\url{https://scikit-learn.org/stable/modules/generated/sklearn.metrics.classification_report.html}}.

\begin{table}[htbp]
	\centering
	\caption{Ablation study for verifying effectiveness of modal fusion on IEMOCAP dataset. T, A and V represent textual, audio and visual modality, respectively.}
	  \begin{tabular}{ccccccr}
			&       &       &       &       &       &  \\
  \cmidrule{2-6}          & \multicolumn{1}{c|}{\multirow{2}[2]{*}{Modality}} & \multicolumn{2}{c|}{ICON} & \multicolumn{2}{c}{MMMIE} &  \\
			& \multicolumn{1}{c|}{} & acc.  & \multicolumn{1}{c|}{F1} & acc.  & F1    &  \\
  \cmidrule{2-6}          & \multicolumn{1}{c|}{T} & 58.3  & \multicolumn{1}{c|}{57.9} & \textbf{61.6} & \textbf{61.0} &  \\
			& \multicolumn{1}{c|}{A} & 50.7  & \multicolumn{1}{c|}{50.9} & 49.2  & 47.5  &  \\
			& \multicolumn{1}{c|}{V} & 41.2  & \multicolumn{1}{c|}{39.8} & 39.6  & 37.0    &  \\
			& \multicolumn{1}{c|}{A+V} & 52.0    & \multicolumn{1}{c|}{51.2} & 51.1  & 50.0    &  \\
			& \multicolumn{1}{c|}{T+A} & 63.8  & \multicolumn{1}{c|}{63.2} & \textbf{66.1} & \textbf{65.3} &  \\
			& \multicolumn{1}{c|}{T+V} & 61.4  & \multicolumn{1}{c|}{61.2} & \textbf{63.9} & \textbf{63.7} &  \\
			& \multicolumn{1}{c|}{T+A+V} & 64.0    & \multicolumn{1}{c|}{63.5} & \textbf{67.8} & \textbf{67.5} &  \\
  \cmidrule{2-6}          &       &       &       &       &       &  \\
	  \end{tabular}%
	\label{ablation_modal}%
  \end{table}%

\textbf{Ablation Study.} In this work, three modules (MI Maximization, MI Minimization and Identity Embedding) are proposed to mine the modality-invariant information, task-related information and contextual information. To verify the effectiveness of different modules and different modalities, a series of ablation experiments are carried out on IEMOCAP. At first, the effectiveness of different modalities is verified, and the results are presented in Table \ref{ablation_modal}. It should be noted that MI maximization module is not applicable, in the case of single modality. Therefore, it can be observed from the Table \ref{ablation_modal} that the model performance under single modal setting is relatively poor. Specifically, the performance corresponding to the visual modality reached the lowest 39.6\%. In general, it is difficult to achieve a satisfactory performance through a single modality in emotion recognition task. It can be also seen from Table \ref{ablation_modal} that the performance can be improved effectively by fusing multiple modalities. Furthermore, the modal based on textual modality achieve the best performance compared to the other two modalities, while the performance of the model based on visual modality is the worst. This phenomenon is consistent with the findings of some previous work \cite{zadeh_tensor_2017} \cite{ghosal2020cosmic}. Just as the conclusions drawn from the previous work, the visual modality in emotion dataset contains a lot of noise and the external state of human beings when expressing emotions is often confusion. These factors together lead to the bad performance of the model on the visual modality and this is also why multimodal fusion is so important for emotion recognition. 

In addition to the ablation study on modalities, the effectiveness of the proposed components also needs to be verified. Therefore, the ablation study on different components is performed to demonstrate the contribution of each components to the model performance and the results are presented in Table \ref{ablation_component}. As can be seen, the MI maximization component has the most significant performance improvement for the model. This means that by maximizing MI between different modal features can mine the modal-invariant information effectively which is the key of multimodal emotion recogntion. Meanwhile, when only two componets are used, by combining the MI Maximization component and MI Minimization component, the performance of this combination is better than that of the other two combinations. This result shows that the optimization of upper bound and lower bound of MI can effectively suppress the noise in the emotion data and mine the modality-invariant information, thus improving the performance and robustness of the model. 


\begin{table}[htbp]
	\centering
	\caption{Ablation study for verifying effectiveness of the proposed components on IEMOCAP dataset. Mmax, Mmin and IE represent the MI Maximization component, MI Minimization component and Identity Embeding component, respectively.}
	  \begin{tabular}{rrrrr}
			&       &       &       &  \\
  \cmidrule{2-4}          & \multicolumn{1}{c}{Measurement} & \multicolumn{2}{c}{Component} &  \\
  \cmidrule{2-4}          &       & \multicolumn{1}{c}{Mmax} & \multicolumn{1}{c}{Mmin} &  \\
  \cmidrule{3-4}          & \multicolumn{1}{c}{acc.} & \multicolumn{1}{c}{64.8} & \multicolumn{1}{c}{64.1} &  \\
			& \multicolumn{1}{c}{F1} & \multicolumn{1}{c}{63.9} & \multicolumn{1}{c}{62.6} &  \\
  \cmidrule{2-4}          &       & \multicolumn{1}{c}{IE} & \multicolumn{1}{c}{Mmax + Mmin} &  \\
  \cmidrule{3-4}          & \multicolumn{1}{c}{acc.} & \multicolumn{1}{c}{63.6} & \multicolumn{1}{c}{67.1} &  \\
			& \multicolumn{1}{c}{F1} & \multicolumn{1}{c}{63.0} & \multicolumn{1}{c}{66.3} &  \\
  \cmidrule{2-4}          &       & \multicolumn{1}{c}{Mmax+IE} & \multicolumn{1}{c}{Mmin+IE} &  \\
  \cmidrule{3-4}          & \multicolumn{1}{c}{acc.} & \multicolumn{1}{c}{66.9} & \multicolumn{1}{c}{64.9} &  \\
			& \multicolumn{1}{c}{F1} & \multicolumn{1}{c}{64.4} & \multicolumn{1}{c}{64.0} &  \\
  \cmidrule{2-4}          &       & \multicolumn{1}{c}{None} & \multicolumn{1}{c}{Mmax+Mmin+IE} &  \\
  \cmidrule{3-4}          & \multicolumn{1}{c}{acc.} & \multicolumn{1}{c}{62.2} & \multicolumn{1}{c}{67.8} &  \\
			& \multicolumn{1}{c}{F1} & \multicolumn{1}{c}{60.7} & \multicolumn{1}{c}{67.5} &  \\
  \cmidrule{2-4}          &       &       &       &  \\
	  \end{tabular}%
	\label{ablation_component}%
  \end{table}%

\subsection{Further Analysis}
\begin{figure}[htb]
	\centering
	\includegraphics[width=90mm, height=50mm]{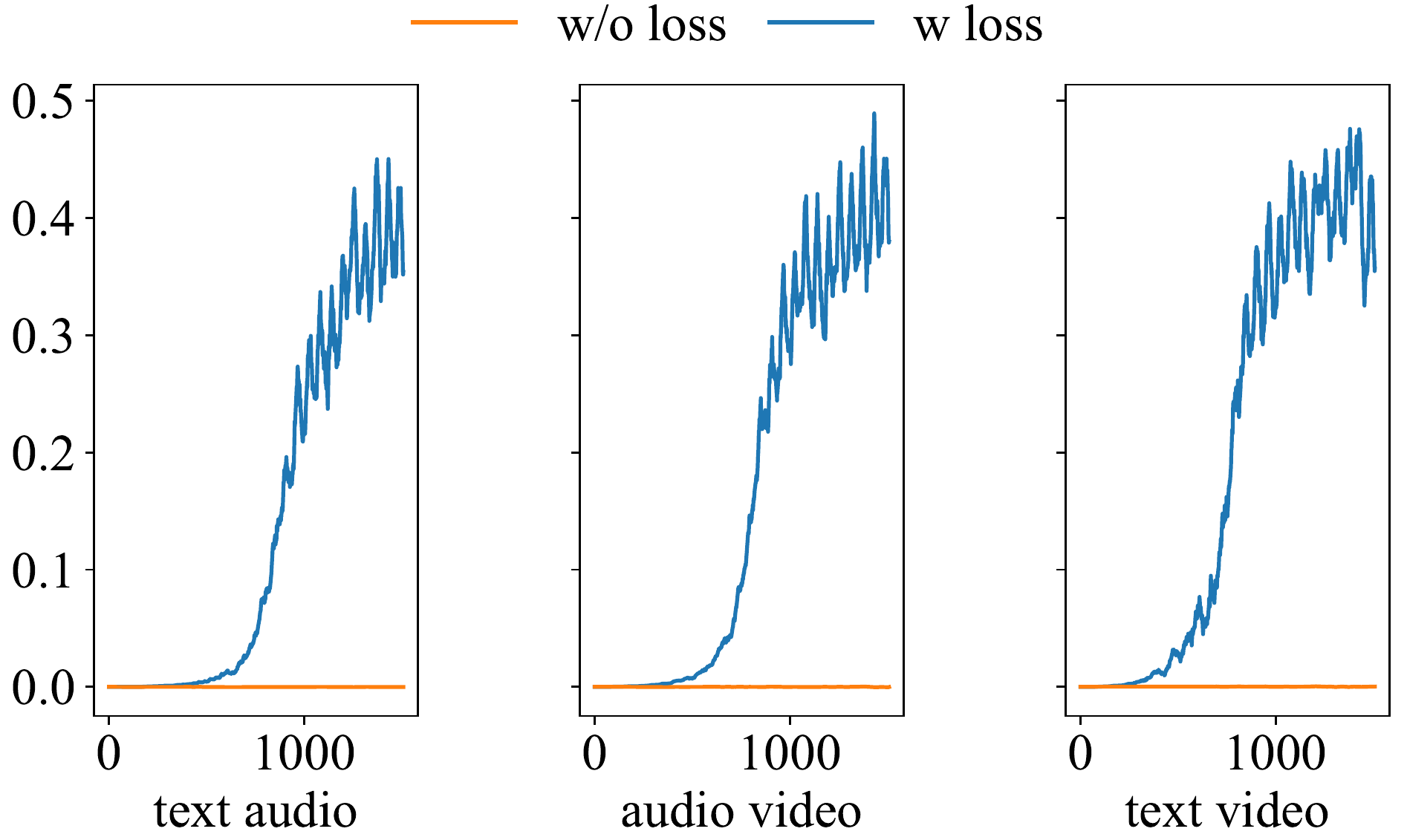}
	\caption{The MI lower bound curve in the training process. The orange curve depicts that the lower bound is not combined in the total optimization objective. The blue curve depicts the lower bound is combined in the total optimization objective. The ``text audio", ``audio video'', and ``text video" represent MI between text and audio, MI between audio and video, and MI between text and video, respectively.}
	\label{fig_mine}
\end{figure}

\begin{figure}[htb]
	\centering
	\includegraphics[width=90mm, height=50mm]{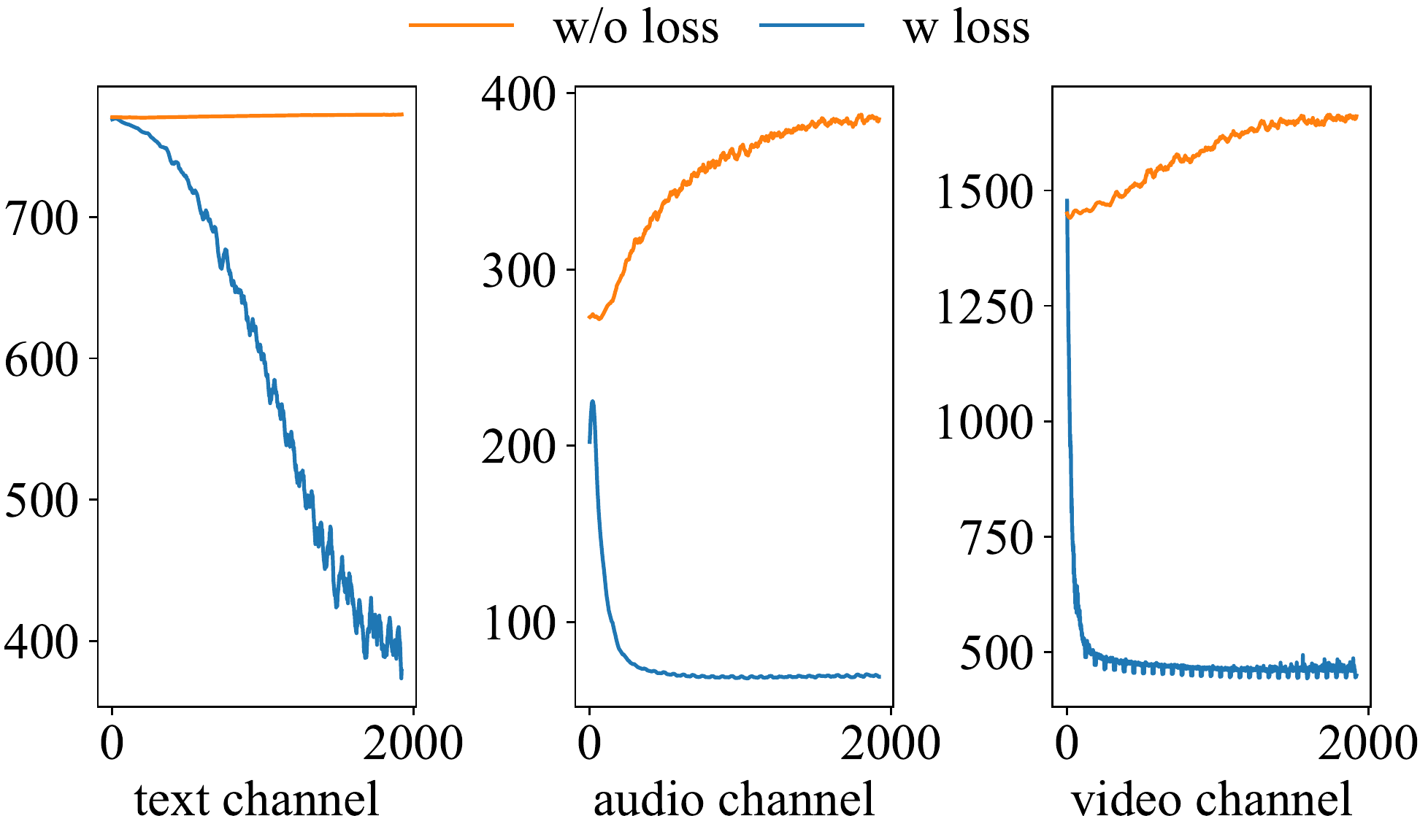}
	\caption{The MI upper bound curve in the training process. The orange curve depicts that the upper bound is not combined in the total optimization objective. The blue curve depicts the upper bound is combined in the total optimization objective. The ``text channel'', ``audio channel'', and ``video channel'' represent the MI in textual modality, audio modality, and video modality, respectively.}
	\label{fig_club}
\end{figure}

\textbf{Convergence Curve.} In order to present the optimization process of MI in more detail, the MI values are recorded. The results are shown in Figure \ref{fig_mine} and \ref{fig_club}. Specifically, Figure \ref{fig_mine} shows the optimization curve of the objective function corresponding to Formula \ref{equa_mi}. The blue curve in this figure represents that the objective function value is added to the total loss and optimized by the back propagation algorithm. In other word, the blue curve reflects the process of MI maximization. Conversely, the orange curve represents that the objective function is not been optimized. This figure indicates that with the training, the lower bound of MI between different modalities can be improved effectively, thereby excavating the invariant information between modalities. Figure \ref{fig_club} shows the optimization curve of the objective function corresponding to Formula \ref{equa_club}. In this figure, the meanings of the blue and orange curves are similar to those in Figure \ref{fig_mine}. It can be found that with the training, the upper bound of MI within modality can be effectively reduced. It should be noted that the orange curve in this figure has a clear upward trend, and it is most pronounced in the audio and image modalities. This trend suggests that if the upper bound of MI is not limited of not optimized, the MI will be improved with training, which usually makes the features of deep layers contain more task-independent information and noise. This result is disastrous for emotion recognition task.

\begin{figure*}[htb]
	\centering
	\includegraphics[width=150mm, height=70mm]{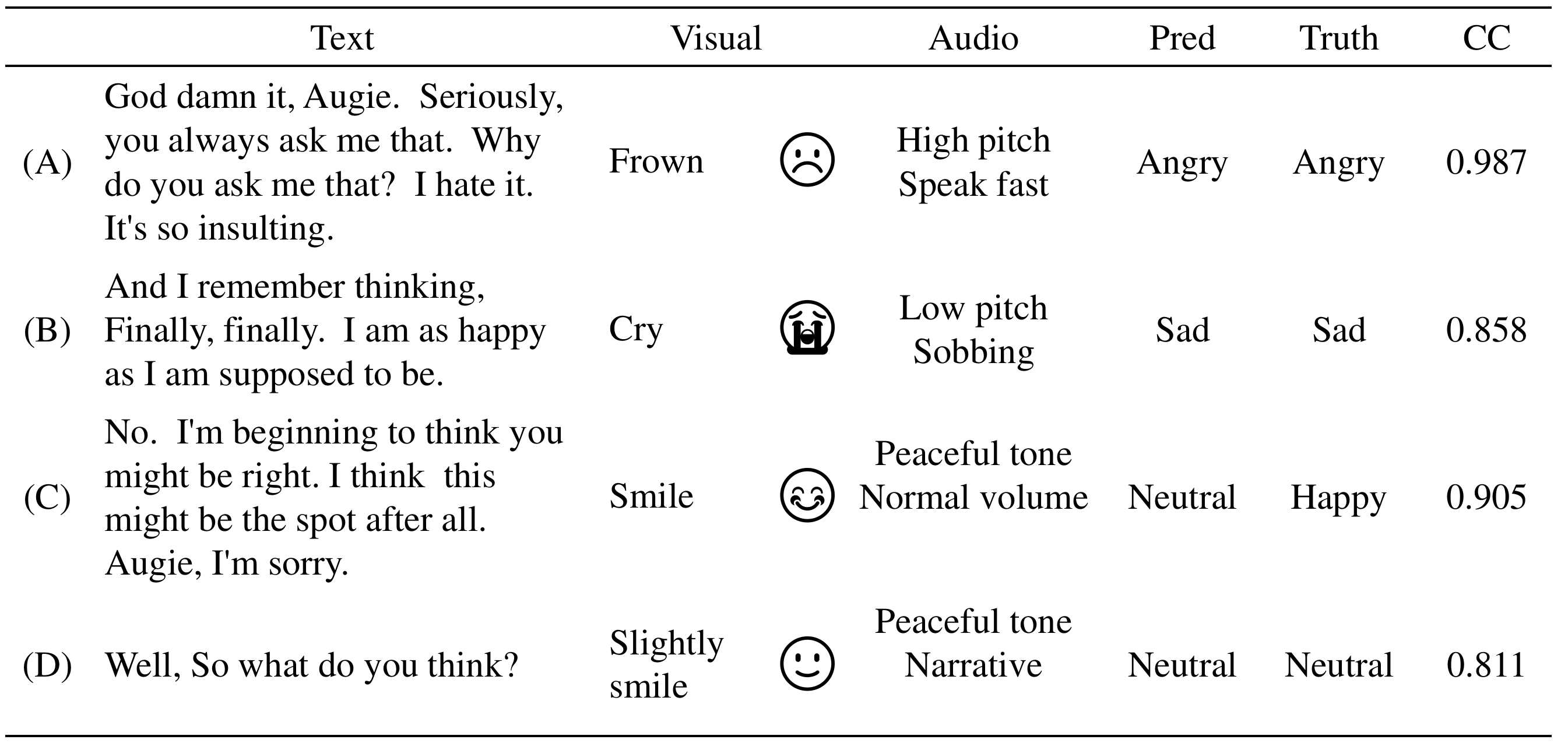}
	\caption{Representative samples with the corresponding predictions and ground truth in the case study. In this figure, the ``Pred'' and ``Truth'' represent the predicted emotion category and the ground truth, respectively. The ``CC'' represents the confidence coefficient of the proposed model output corresponding to the predicted emotion category.}
	\label{fig_case}
\end{figure*}
\textbf{Case Study.} Besides the aforementioned analysis, some data randomly extracted from the IEMOCAP dataset are also used to specifically show the recognition results of the model. As shown in Figure \ref{fig_case}, four samples are selected for analysis. Since the visual and audio data are inconvenient to present directly in the manuscript, the two modalities are illustrated literally. In case (A), the text modality provides some obvious emotional cues, such as words like ``hate'' and ``insulting''. Similarly, visual and audio modalities also provide obvious information related to emotion, thus resulting in a high confidence recognition result. In case (B), although text modality does not provide obvious emotional cues, visual and audio modalities provide. This case demonstrates that when one modality can not provide effective information, the recognition accuracy can be guaranteed by combining it with other modalities and utilizing the common feature hidden among these modalities. In case (C), some words with ambiguous meanings (such as ``sorry'') appear in the text modalities, which may lead the model to output wrong results. Therefore, although the visual and audio modalities provide information related to the ground truth, the model ends up producing erroneous results, suggesting that the proposed model may potentially pay more attention to the information of the text modality. We analyze that this is due to the multi attention layers used in our network and the higher robustness and less noise of the text modality compared to other modalities in the data set. In case (D), all modalities do not have strong emotional overtones, so the model can accurately recognize the neutral category. This case is similar to A. Multiple modalities are strongly correlated with ground truth. This type of data is the easiest for the model to identify. In general, the proposed model can effectively use the modal invariant information to improve the recognition results, but when encountering confusing words in text modalities, there will be bad cases, which will be what we need to improve in the future.

\section{CONCLUSION}
This work is motivated by how to construct a robust multimodal representation to bridge the heterogeneity gap between different multimodal information and how to model the contextual dynamics in a conversation efficiently for multimodal sentiment analysis. Different from the previous works focusing on geometric manipulation in feature space or combining features from different utterances by using algorithms such as attention mechanism or LSTM, we propose a novel framework, namely MMMIE, that cleverly combines mutual information maximization and minimization. Through maximizing the mutual information between different modal pairs, the cross-modal and intra-modal dynamics are modeled throughout the information flow. By minimizing the mutual information between input data and corresponding features, the redundant information can be filtered out to improve the robustness of the model. In addition, IE is proposed to prompt the model to perceive the contextual information from the shallow layer to the deep layer. Comprehensive experiments are conducted on two public datasets and the results demonstrate the effectiveness of the proposed model. This work will inspire creativity in multimodal representation learning and multimodal sentiment analysis in the future.
\section*{Acknowledgments}
The work was supported by the National Natural Science Foundation of China under Grant no. 61876209 and the National Key Research and Development Program of China under Grant no. 2017YFC1501301.

\bibliography{mybibfile}

\begin{thebibliography}{10}
\expandafter\ifx\csname url\endcsname\relax
  \def\url#1{\texttt{#1}}\fi
\expandafter\ifx\csname urlprefix\endcsname\relax\def\urlprefix{URL }\fi
\expandafter\ifx\csname href\endcsname\relax
  \def\href#1#2{#2} \def\path#1{#1}\fi

\bibitem{tsai2018learning}
Y.-H.~H. Tsai, P.~P. Liang, A.~Zadeh, L.-P. Morency, R.~Salakhutdinov, Learning
  factorized multimodal representations, arXiv preprint arXiv:1806.06176.

\bibitem{zadeh_multi_2018}
A.~Zadeh, P.~P. Liang, S.~Poria, P.~Vij, E.~Cambria, L.-P. Morency,
  Multi-attention recurrent network for human communication comprehension,
  Proceedings of the AAAI Conference on Artificial Intelligence 32~(1).

\bibitem{tao2009affective}
J.~Tao, T.~Tan, Affective information processing, Springer London, 2009.

\bibitem{zadeh_tensor_2017}
A.~Zadeh, M.~Chen, S.~Poria, E.~Cambria, L.-P. Morency, Tensor fusion network
  for multimodal sentiment analysis, arXiv preprint arXiv:1707.07250.

\bibitem{liu_efficient_2018}
Z.~Liu, Y.~Shen, V.~B. Lakshminarasimhan, P.~P. Liang, A.~Zadeh, L.-P. Morency,
  Efficient low-rank multimodal fusion with modality-specific factors, arXiv
  preprint arXiv:1806.00064.

\bibitem{dumpala_audio_2019}
S.~H. Dumpala, I.~Sheikh, R.~Chakraborty, S.~K. Kopparapu, Audio-visual fusion
  for sentiment classification using cross-modal autoencoder, in: Proc. Neural
  Inf. Process. Syst.(NIPS), 2019, pp. 1--4.

\bibitem{sahu_dynamic_2019}
G.~Sahu, O.~Vechtomova, Dynamic fusion for multimodal data, arXiv preprint
  arXiv:1911.03821.

\bibitem{hazarika2018conversational}
D.~Hazarika, S.~Poria, A.~Zadeh, E.~Cambria, L.-P. Morency, R.~Zimmermann,
  Conversational memory network for emotion recognition in dyadic dialogue
  videos, in: Proceedings of the conference. Association for Computational
  Linguistics. North American Chapter. Meeting, Vol. 2018, NIH Public Access,
  2018, p. 2122.

\bibitem{majumder2019dialoguernn}
N.~Majumder, S.~Poria, D.~Hazarika, R.~Mihalcea, A.~Gelbukh, E.~Cambria,
  Dialoguernn: An attentive rnn for emotion detection in conversations, in:
  Proceedings of the AAAI Conference on Artificial Intelligence, Vol.~33, 2019,
  pp. 6818--6825.

\bibitem{ghosal2020cosmic}
D.~Ghosal, N.~Majumder, A.~Gelbukh, R.~Mihalcea, S.~Poria, Cosmic: Commonsense
  knowledge for emotion identification in conversations, arXiv preprint
  arXiv:2010.02795.

\bibitem{belghazi2018mine}
M.~I. Belghazi, A.~Baratin, S.~Rajeshwar, S.~Ozair, Y.~Bengio, A.~Courville,
  D.~Hjelm, Mutual information neural estimation, in: Proceedings of the 35th
  International Conference on Machine Learning, Vol.~80, PMLR, 2018, pp.
  531--540.

\bibitem{cheng2020club}
P.~Cheng, W.~Hao, S.~Dai, J.~Liu, Z.~Gan, L.~Carin, Club: A contrastive
  log-ratio upper bound of mutual information, in: International Conference on
  Machine Learning, PMLR, 2020, pp. 1779--1788.

\bibitem{vaswani2017attention}
A.~Vaswani, N.~Shazeer, N.~Parmar, J.~Uszkoreit, L.~Jones, A.~N. Gomez,
  L.~Kaiser, I.~Polosukhin, Attention is all you need, in: Advances in Neural
  Information Processing Systems, Vol.~30, 2017.

\bibitem{hochreiter1997long}
S.~Hochreiter, J.~Schmidhuber, Long short-term memory, Neural computation 9~(8)
  (1997) 1735--1780.

\bibitem{busso_iemocap_2008}
C.~Busso, M.~Bulut, C.-C. Lee, A.~Kazemzadeh, E.~Mower, S.~Kim, J.~N. Chang,
  S.~Lee, S.~S. Narayanan, Iemocap: Interactive emotional dyadic motion capture
  database, Language resources and evaluation 42~(4) (2008) 335.

\bibitem{poria2018meld}
S.~Poria, D.~Hazarika, N.~Majumder, G.~Naik, E.~Cambria, R.~Mihalcea, Meld: A
  multimodal multi-party dataset for emotion recognition in conversations,
  arXiv preprint arXiv:1810.02508.

\bibitem{ngiam_multimodal_2011}
J.~Ngiam, A.~Khosla, M.~Kim, J.~Nam, H.~Lee, A.~Y. Ng, Multimodal deep
  learning, in: Proceedings of the 28th International Conference on Machine
  Learning (ICML-11), 2011, pp. 689--696.

\bibitem{lee_sparse_2008}
H.~Lee, C.~Ekanadham, A.~Ng, Sparse deep belief net model for visual area v2,
  in: Advances in Neural Information Processing Systems, 2008, pp. 873--880.

\bibitem{wang_deep_2015}
D.~Wang, P.~Cui, M.~Ou, W.~Zhu, Deep multimodal hashing with orthogonal
  regularization, in: Proceedings of the 24th International Conference on
  Artificial Intelligence, 2015, pp. 2291--2297.

\bibitem{andrew_deep_2013}
G.~Andrew, R.~Arora, J.~Bilmes, K.~Livescu, Deep canonical correlation
  analysis, in: International conference on machine learning, 2013, pp.
  1247--1255.

\bibitem{liu_multimodal_2019}
W.~Liu, J.-L. Qiu, W.-L. Zheng, B.-L. Lu, Multimodal emotion recognition using
  deep canonical correlation analysis, arXiv preprint arXiv:1908.05349.

\bibitem{han2021improving}
W.~Han, H.~Chen, S.~Poria, Improving multimodal fusion with hierarchical mutual
  information maximization for multimodal sentiment analysis, arXiv preprint
  arXiv:2109.00412.

\bibitem{tsai_multimodal_2019}
Y.-H.~H. Tsai, S.~Bai, P.~P. Liang, J.~Z. Kolter, L.-P. Morency,
  R.~Salakhutdinov, Multimodal transformer for unaligned multimodal language
  sequences, in: Proceedings of the conference. Association for Computational
  Linguistics. Meeting, 2019, pp. 6558--6569.

\bibitem{liu2021contrastive}
Y.~Liu, Q.~Fan, S.~Zhang, H.~Dong, T.~Funkhouser, L.~Yi, Contrastive multimodal
  fusion with tupleinfonce, in: Proceedings of the IEEE/CVF International
  Conference on Computer Vision, 2021, pp. 754--763.

\bibitem{hazarika2018icon}
D.~Hazarika, S.~Poria, R.~Mihalcea, E.~Cambria, R.~Zimmermann, Icon:
  Interactive conversational memory network for multimodal emotion detection,
  in: Proceedings of the 2018 conference on empirical methods in natural
  language processing, 2018, pp. 2594--2604.

\bibitem{ghosal2019dialoguegcn}
D.~Ghosal, N.~Majumder, S.~Poria, N.~Chhaya, A.~Gelbukh, Dialoguegcn: A graph
  convolutional neural network for emotion recognition in conversation, arXiv
  preprint arXiv:1908.11540.

\bibitem{shen2021directed}
W.~Shen, S.~Wu, Y.~Yang, X.~Quan, Directed acyclic graph network for
  conversational emotion recognition, arXiv preprint arXiv:2105.12907.

\bibitem{devlin2018bert}
J.~Devlin, M.-W. Chang, K.~Lee, K.~Toutanova, Bert: Pre-training of deep
  bidirectional transformers for language understanding, arXiv preprint
  arXiv:1810.04805.

\bibitem{he_deep_2016}
K.~He, X.~Zhang, S.~Ren, J.~Sun, Deep residual learning for image recognition,
  in: Proceedings of the IEEE conference on computer vision and pattern
  recognition, 2016, pp. 770--778.

\bibitem{baevski2020wav2vec}
A.~Baevski, H.~Zhou, A.~Mohamed, M.~Auli, wav2vec 2.0: A framework for
  self-supervised learning of speech representations, arXiv preprint
  arXiv:2006.11477.

\bibitem{liu2019multi}
X.~Liu, P.~He, W.~Chen, J.~Gao, Multi-task deep neural networks for natural
  language understanding, arXiv preprint arXiv:1901.11504.

\bibitem{doddington1985speaker}
G.~Doddington, Speaker recognition—identifying people by their voices,
  Proceedings of the IEEE 73~(11) (1985) 1651--1664.
\newblock \href {http://dx.doi.org/10.1109/PROC.1985.13345}
  {\path{doi:10.1109/PROC.1985.13345}}.

\bibitem{zhuang2019investigation}
B.~Zhuang, W.~Wang, T.~Shinozaki, Investigation of attention-based multimodal
  fusion and maximum mutual information objective for dstc7 track3, in: DSTC7
  at AAAI2019 workshop, 2019.

\bibitem{datta2017multimodal}
D.~Datta, S.~Varma, S.~K. Singh, et~al., Multimodal retrieval using mutual
  information based textual query reformulation, Expert Systems with
  Applications 68 (2017) 81--92.

\bibitem{wu2020graph}
T.~Wu, H.~Ren, P.~Li, J.~Leskovec, Graph information bottleneck, arXiv preprint
  arXiv:2010.12811.

\bibitem{zhong2019knowledge}
P.~Zhong, D.~Wang, C.~Miao,
  \href{https://aclanthology.org/D19-1016}{Knowledge-enriched transformer for
  emotion detection in textual conversations}, in: Proceedings of the 2019
  Conference on Empirical Methods in Natural Language Processing and the 9th
  International Joint Conference on Natural Language Processing (EMNLP-IJCNLP),
  Association for Computational Linguistics, Hong Kong, China, 2019, pp.
  165--176.
\newblock \href {http://dx.doi.org/10.18653/v1/D19-1016}
  {\path{doi:10.18653/v1/D19-1016}}.
\newline\urlprefix\url{https://aclanthology.org/D19-1016}

\bibitem{li2022bieru}
W.~Li, W.~Shao, S.~Ji, E.~Cambria, Bieru: Bidirectional emotional recurrent
  unit for conversational sentiment analysis, Neurocomputing 467 (2022) 73--82.

\bibitem{li2020hitrans}
J.~Li, D.~Ji, F.~Li, M.~Zhang, Y.~Liu, Hitrans: A transformer-based context-and
  speaker-sensitive model for emotion detection in conversations, in:
  Proceedings of the 28th International Conference on Computational
  Linguistics, 2020, pp. 4190--4200.

\bibitem{Ishiwatari2020RGAT}
T.~Ishiwatari, Y.~Yasuda, T.~Miyazaki, J.~Goto,
  \href{https://aclanthology.org/2020.emnlp-main.597}{Relation-aware graph
  attention networks with relational position encodings for emotion recognition
  in conversations}, in: Proceedings of the 2020 Conference on Empirical
  Methods in Natural Language Processing (EMNLP), Association for Computational
  Linguistics, Online, 2020, pp. 7360--7370.
\newblock \href {http://dx.doi.org/10.18653/v1/2020.emnlp-main.597}
  {\path{doi:10.18653/v1/2020.emnlp-main.597}}.
\newline\urlprefix\url{https://aclanthology.org/2020.emnlp-main.597}

\bibitem{poria2017context}
S.~Poria, E.~Cambria, D.~Hazarika, N.~Majumder, A.~Zadeh, L.-P. Morency,
  Context-dependent sentiment analysis in user-generated videos, in:
  Proceedings of the 55th annual meeting of the association for computational
  linguistics (volume 1: Long papers), 2017, pp. 873--883.

\bibitem{zadeh_memory_2018}
A.~Zadeh, P.~P. Liang, N.~Mazumder, S.~Poria, E.~Cambria, L.-P. Morency, Memory
  fusion network for multi-view sequential learning, in: Proceedings of the
  AAAI Conference on Artificial Intelligence, Vol.~32, 2018.

\end{thebibliography}
\end{document}